\documentclass[lettersize,journal]{IEEEtran}
\usepackage{amsmath,amsfonts}
\usepackage{algorithmic}
\usepackage{algorithm}
\usepackage{array}
\usepackage[caption=false,font=normalsize,labelfont=sf,textfont=sf]{subfig}
\usepackage{colortbl}
\usepackage{textcomp}
\usepackage{stfloats}
\usepackage{url}
\usepackage{verbatim}
\usepackage{multicol}
\usepackage{multirow}
\usepackage{graphicx}
\usepackage{textcomp,booktabs}
\usepackage{cite}
\begin{document}

\title{Vertical Federated Learning: Challenges, Methodologies and Experiments}

\author{Kang Wei,~\IEEEmembership{Graduate Student Member,~IEEE,}
        Jun Li,~\IEEEmembership{Senior Member,~IEEE,}
        Chuan Ma,~\IEEEmembership{Member,~IEEE,}

        Ming Ding,~\IEEEmembership{Senior Member,~IEEE,}
        Sha Wei,
        Fan Wu,~\IEEEmembership{Member,~IEEE}
        and Guihai Chen,~\IEEEmembership{Senior Member,~IEEE}
\thanks{Kang~Wei, Jun~Li and Chuan~Ma are with School of Electrical and Optical Engineering, Nanjing University of Science and Technology, Nanjing 210094, China. Chuan Ma is also with Key Laboratory of Computer Network and Information Integration (Southeast University), Ministry of Education (e-mail: \{kang.wei; jun.li; chuan.ma\}@njust.edu.cn).}
\thanks{Ming~Ding is with Data61, CSIRO, Sydney, NSW 2015, Australia (e-mail: ming.ding@data61.csiro.au).}
\thanks{Sha Wei is with China Academy of Information and Communication Technology (CAICT), Beijing 100191, China (e-mail: weisha@caict.ac.cn).}
\thanks{Fan Wu is with the Shanghai Key Laboratory of Scalable Computing
and Systems, Department of Computer Science and Engineering, Shanghai
Jiao Tong University, Shanghai 200240, China (e-mail: fwu@cs.sjtu.edu.cn).}
\thanks{Guihai Chen is with the State Key Laboratory for Novel Software Technology, Nanjing University, Nanjing 210023, China (e-mail: gchen@nju.edu.cn).}}

\markboth{Journal of \LaTeX\ Class Files,~Vol.~14, No.~8, August~2021}%
{Shell \MakeLowercase{\textit{et al.}}: A Sample Article Using IEEEtran.cls for IEEE Journals}


\maketitle

\begin{abstract}
Recently, federated learning (FL) has emerged as a promising distributed machine learning (ML) technology, owing to the advancing computational and sensing capacities of end-user devices, as well as the increasing concerns on users' privacy.
As a special architecture in FL, vertical FL (VFL) is capable of constructing a hyper ML model by embracing sub-models from different types of clients.
These sub-models are trained locally by vertically partitioned data with distinct attributes.
Therefore, the design of VFL is fundamentally different from that of conventional FL, raising new and unique research issues.
In this paper, we aim to discuss key challenges in VFL with effective solutions, and conduct experiments on real-life datasets to shed light on these issues.
Specifically, we first propose a general framework on VFL, and highlight the key differences between VFL and conventional FL.
Then, we discuss research challenges rooted in VFL systems from four aspects, i.e., security and privacy risks, expensive computation and communication costs, possible structural damage caused by model splitting, and system heterogeneity.
Afterwards, we develop solutions to addressing the aforementioned challenges, and conduct extensive experiments to showcase the effectiveness of the proposed solutions.
\end{abstract}


\section{Introduction}
\IEEEPARstart{W}{ith} the emergence of end-user devices, equipped with various sensors and increasingly powerful hardwares, huge amounts of data are generated through day-to-day usage in recent years~\cite{Nguyen2021Federated}.
In a concurrent development, machine learning (ML) has revolutionized the ways that information is extracted with ground breaking successes in areas such as computer vision, natural language processing, voice recognition, etc.
Therefore, there is a high demand for harnessing the rich data provided by distributed devices or data owners to enrich ML models.
To address this key issue as well as ensuring data privacy and security, many distributed learning systems have been widely exploited to train data models~\cite{Wei2020Fed}.

As a new distributed learning diagram, federated learning (FL) has drawn a lot of attentions by involving training statistical models over remote devices or siloed data centers, such as mobile phones or hospitals, while keeping data locally~\cite{Ma2019FL}.
In terms of data partitions, FL can be categorized into three paradigms, i.e., horizontal, vertical and transfer types.
Most of existing works focus on horizontal FL (HFL) that requires all participants possess the same attribute space but different sample spaces.
It is suitable for mobile devices and usually involves billions of devices with heterogeneous resources under a complex distributed network~\cite{Yu2021Toward}.
However, application cases of HFL are limited due to practical reasons, such as the confidentiality among companies with a same competing interest.
Vertical FL (VFL), on the other hand, can avoid this issue by promoting collaborations among non-competing organizations/entities with vertically partitioned data, e.g., a collaboration among a bank, an insurance company, and an e-commerce platform to learn users' living/shopping behaviors.

The success of VFL is primarily owing to its application scenarios~\cite{Cheng2020Federated,Fu2021VF2Boost}, such as Fedlearner\footnote{https://github.com/bytedance/fedlearner} in bytedance and Angel PowerFL\footnote{https://data.qq.com/powerfl} in Tencent.
Indeed, compared with independent training or HFL, VFL can exploit more/deeper attribute dimensions, and obtain a better learning model.
One of its typical usage scenarios, which is the focus of this paper, is that a few participants collaboratively train a model with distributed attributes but labels of these attributes are owned by only one participant.
For example, a car insurance company with limited user attributes might want to improve the risk evaluation model by incorporating more attributes from other businesses, e.g., a bank, a taxation office, etc.
The role of the other participants is simply providing additional feature information without directly disclosing their data, and in return, obtain financial and/or reputational rewards.
In addition, VFL also requires consistently fewer resources from the participating client compared with data sharing directly, enabling lightweight and scalable distributed training solutions.
However, up to now, few efforts have been spent on investigating the core challenges and methodologies of this ML framework.

Inspired by this research gap, in this article, we investigate the potential challenges as well as methodologies for VFL systems.
Specifically, we first propose a general framework on VFL in details.
Then, we clarify the key differences between VFL and HFL, including data characteristics of participants, exchanged messages and model structures.
Furthermore, we discuss potential unique challenges and possible solutions.
Finally, we have conducted three typical experiments, e.g., differential privacy (DP) aided VFL, compression empowered communication efficiency, and splitting design for resource allocation.
The remainder of this article is organized as follows.
The next section introduces the basic model of VFL and key differences between VFL and HFL.
Then we illustrate challenges in developing VFL in Section~\ref{sec:challenges}, and provide probable solutions for these challenges in Section~\ref{sec:solutions}.
In Section~\ref{sec:experi_results}, three typical experiments have been conducted to evaluate the proposed solutions.
Finally, conclusions are drawn in Section~\ref{sec:conclusions}.
\begin{figure*}[!t]
\centering
\includegraphics[width=6.5in]{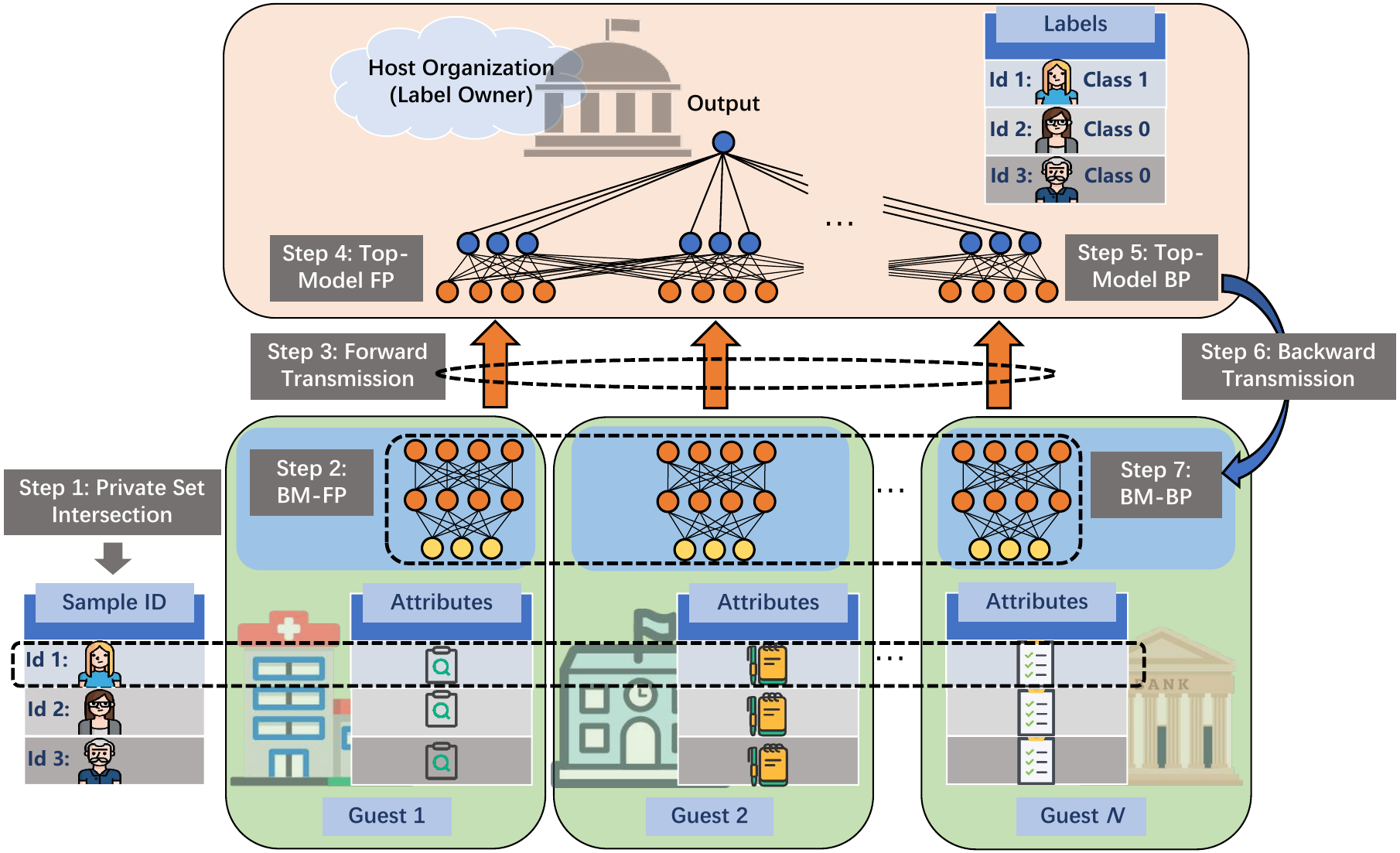}
\caption{A diagram for Vertical FL. The classic workflow includes following seven steps: 1) private set intersection; 2) bottom model forward propagation (BM-FP); 3) forward transmission; 4) top model forward propagation (TM-FP); 5) top model backward propagation (TM-BP); 6) backward transmission; 7) bottom model backward propagation (BM-BP). The host is the label owner and the guest is the attribute owner.}
\label{fig:VFL_systems}
\end{figure*}
\section{Background}
\subsection{Frameworks for VFL}
The key idea of VFL is to enhance the learning model by utilizing the distributed data with various attributes.
Hence, it is designed for the vertically partitioned data where participants' data share the same sample space with different attribute spaces.
As shown in Fig.~\ref{fig:VFL_systems}, a general VFL process for each learning epoch includes the following seven key steps:
\begin{itemize}
\item[$\bullet$]\textbf{Private set intersection.}
Before model training, the framework needs to find the common IDs served by all participants (i.e., guest organization and host organization) to align the training data samples, which is called private set intersection (PSI) or secure entity alignment~\cite{Lu2020Multi}.
PSI is an interactive cryptographic protocol which allows multiple participants to find out the common IDs and nothing else.
Widely adopted PSI techniques include naive hashing, oblivious polynomial evaluation, and oblivious transfer~\cite{Lu2020Multi}.
\item[$\bullet$]\textbf{Bottom model forward propagation.}
After determining the aligned data samples among all participants, each participant will complete a forward propagation process using local data based on its bottom model.
This forward propagation process is similar to the conventional training except calculating the loss value.
\item[$\bullet$]\textbf{Forward output transmission.}
Each participant needs to transmit its forward output to the label owner.
Intuitively speaking, the forward output contains intermediate results of local neural networks, which transforms the original attributes into features.
Such a transmission process may divulge participants' privacy information.
Hence, advanced privacy preserving methods should be exploited to address this potential risk but will may incur additional communication costs and computation complexities.
\item[$\bullet$]\textbf{Top model forward propagation.}
The label owner uses the collected outputs from all participants to calculate the loss function value based on the top model and labels.
\item[$\bullet$]\textbf{Top model backward propagation.}
The label owner performs backward propagation and computes two gradients for: 1) model parameters of the top model; and 2) forward outputs from every participant.
Using the gradients of the top model, the label owner can calculate the average gradients for each batch and update its model.
\item[$\bullet$]\textbf{Backward output transmission.}
The gradients of forward outputs are sent back to every participant.
It can be noticed the required transmission bits are usually much smaller than the ones in \textbf{Step 2}, because they are gradients instead of intermediate outputs.
\item[$\bullet$]\textbf{Bottom model backward propagation.}
Each participant calculates the gradients of its bottom model parameters based on the local data and gradients of the forward outputs from the label owner, and then updates its bottom model.
\end{itemize}
After introducing the VFL process for each learning epoch, we will discuss the differences between VFL and HFL in the following subsection.

\begin{table*}[hbt]
\centering
\caption{Key differences between VFL and HFL}
\label{tab:key_differences}
\begin{tabular}{ccccc}
\toprule
Frameworks& \multicolumn{2}{c}{Data characteristics of participants}&Exchanged messages& Model structures \\
\midrule
VFL&Same sample space&Different attribute spaces&Intermediate outputs and its gradients&Flexible and local secret\\
\hline
HFL&Different sample spaces &Same attribute space&Global and local model parameters&Fixed and consistent\\
\toprule
\end{tabular}
\end{table*}

\subsection{Key Differences between VFL and HFL}
In this subsection, the differences between VFL and HFL can be summarized in three aspects, i.e., data characteristics, exchanged messages and model structures, which is shown as follows:
\begin{itemize}
\item[$\bullet$] VFL systems require that all participants possess the same sample space and different attribute spaces, e.g., the collaboration between a bank and an e-commerce platform, but HFL can only be conducted among participants with the same attribute space and different sample spaces.
    This inherent difference between VFL and HFL leads to distinctively different neural network structures.
    For a concrete example, there could be hundreds or thousands of participants in HFL, while the number of participants in VFL is usually less than five.
    As a result, hot topics of HFL, such as participant selection, do not apply to VFL and unique problems such as PSI protocols can only be studied in the framework of VFL.
\item[$\bullet$] The main characteristic of HFL is that each participant maintains a local model and receives the global model periodically.
    However, each participant in VFL possesses a part of a full model, and all participants should finish a training process part by part.
    Therefore, exchanged messages among participants in VFL are the intermediate outputs (learning representations) of the local data based on bottom models and their gradients, instead of local model parameters or updates in HFL.
\item[$\bullet$] The communication cost, and security and privacy risks are much different from the ones in the HFL because of their different structures.
    For example, the bottom model structure in VFL for each participant is flexible and a local secret (unknown to others), which is determined by the splitting methods.
    In this case, the splitting methods after careful design, need to be suitable for specific models, heterogeneous attribute space, and communication and computing resources.
\end{itemize}

To demonstrate key differences between VFL and HFL, we summarize main ideas in Tab.~\ref{tab:key_differences}.
In the following section, we will introduce the challenges of VFL in details.
\section{Core Challenges}\label{sec:challenges}
In this section, we will discuss four of the core challenges associated with solving the distributed learning problem for VFL.
These raised challenges make the VFL design distinct from other classical problems, such as the centralized learning or HFL.
\subsection{Security and Privacy Risks}
Researches on the privacy and security risks for existing VFL models are insufficient.
In VFL, participants need to obtain the coincident sample space, thus the membership inference attack from other participants may be redundant.
In addition, in VFL, the adversarial participant only controls part of the federated model, which cannot run independently, and only has access to the gradients of the incomplete model.
However, via analysing exchanged messages, i.e., intermediate outputs and the gradients, participants can infer the clients' attributes from other participants, such as the label inference attack~\cite{Fu2022Label} and private data leakage~\cite{Jin2021CAFE}.
Protecting the privacy of the labels owned by each participant should be a fundamental requirement in VFL, as the labels might be highly sensitive, e.g., whether a person has a certain type of disease.
In some special cases, the gradients from the server can also directly leak the label information.
Furthermore, it is also been proven that recovering batch data from the shared gradients is available.
We can also note the attacking methods varies in different data types.
For example, tabular data usually needs embedding before learning, and it is difficult obtain the private information from embedded attributes.
It is still a research question whether adversaries can take advantage of the exchanged messages in VFL to recover raw attribute values.

Existing privacy preserving methods including DP, secure multiparty computation (SMC), homomorphic encryption (HE), and their hybrid methods, have been widely adopted in HFL, but few ones have been explored in VFL.
In addition, these methods further need a well designed tradeoff in terms of the model performance, the privacy and security level, and the system efficiency.
An unique challenge for VFL is that splitting methods will determine the preprocess of secret attributes, i.e., forward outputs of the bottom model, and then affect the privacy and security level.
For the simple models, such as logistic regression (LR)~\cite{Yang2019Federated} and kernel models~\cite{Gu2020Federated}, the splitting design is straightforward, but privacy preserving methods are considerable challenging due to the insecure linear process for the input.
If complex neural networks are considered, e.g., the convolutional network, as the bottom model, the intermediate outputs will expose less private information.
This is due to a fact that a complex nonlinear function process can naturally enhance the privacy of attributes.
Understanding and balancing these tradeoffs among privacy protection, operating efficiency and model performance, both theoretically and empirically, is a considerable challenge in realizing private VFL systems.
\subsection{High-cost of Computation and Communication}
Although the raw data is not explicitly shared in the federated setting, resource-limited communication networks are still critical bottlenecks in both HFL and VFL systems~\cite{Abhishek2019Detailed}.
Specifically, in VFL, the total computation and communication cost is proportional to the training dataset size.
In other words, the widely adopted batch computation method in HFL cannot be applied to VFL.
When facing a huge amount of data, e.g., billions of advertising data, communication and local computation may be unmatched by many orders of magnitude due to limited resources, such as hardware capacity, bandwidth, and power.
To fit a limited resources condition in VFL network, it is therefore important to develop computation and communication-efficient methods that reduce the computation complexity and iteratively send the messages or model updates as part of the training process, respectively.
To further reduce computation and communication cost in such a setting, three key aspects can considered 1) pruning the neural network, 2) reducing the size of the transmitted messages by smart compression methods at each round and 3) selecting the proper splitting position.
\subsection{Structural Damage of Model Splitting}
With various datasets, the best corresponding training model is usually carefully designed, but the model splitting process may destroy its specific structure in VFL.
For example, some classic recommendation models for advertising data, e.g., Wide$\&$Deep and deepFM, need to calculate cosine similarities between user and item attributes to explore their relevances.
However, disparate attribute values and privacy concerns make it difficult to meet this requirement.
Furthermore, for the much complex but efficient neural network models that contain recurrent and attention layers, e.g., transformer, the splitting design becomes challenging in realizing privacy preserving and communication efficiency.


\subsection{System Heterogeneity}
The variability of each participant in storage, hardware (CPU/GPU and memory), network connectivity (5G and wifi), and power (battery level) will incur the system heterogeneity.
Additionally, the network size and systems-related constraints on each participant typically result in asynchronous updates~\cite{Zhang2021Secure}, i.e., a part of participants being active at once.
For example, if one participant possess amounts of attributes but limited transmission capacity, computation frequencies or memory size, it will be difficult to complete the bottom model forward propagation.
These system-level characteristics dramatically pose challenges, such as straggler mitigation and fault tolerance.
Therefore, developed VFL methods must satisfy: 1) anticipating a low amount of participation 2) tolerating heterogeneous hardware, and 3) being robust enough to dropped exchanged messages in one iteration.
\section{Methodologies and Possible Solutions}\label{sec:solutions}
In this section, we will analyse the available techniques to address the aforementioned challenges, and provide some possible solutions.
\subsection{Privacy Preserving Frameworks}\label{subsec:priv_pre}
In a VFL system, security and privacy issues usually focus on the exchanged messages, whatever training or serving.
Therefore, it is significant to protect these private information as well as maximizing the training utility.
Some popular techniques, such as DP, SMPC and HE, can be adopted for this issue but usually induce other damages to the learning system.
\begin{itemize}
\item[$\bullet$]\textbf{Differential privacy.} DP is a popular research direction to enhance the privacy level based on theoretical aspects~\cite{Abuadbba2020Sun}.
    Along with other privacy enhancing techniques, such as data shuffle, driving a tight bound for the privacy loss is beneficial to obtaining a better tarde-off between utility and privacy.
    Specifically, for the high efficiency requirement, against the inference attack, DP will be a good selection via perturbing the backward output or local bottom networks.
\item[$\bullet$]\textbf{Secure multi-party computing.} SMPC is used to design various privacy-preserving protocols by involving encryption and protocol design, which is also worth investigating for VFL systems, especially for the limited communication and computation resources.
    Compared with DP, this mechanism can maintain the original utility, but incurs a higher complexity due to cipher inflation and more information interaction.
\item[$\bullet$]\textbf{Homomorphic encryption.} The high complexity is a key bottleneck to consider HE when developing security and privacy methods for federated networks.
    Only the addition and multiplication can be achieved by HE, thus the nonlinear operation requires some well-designed approximate functions and performance losses are inevitable.
    Fortunately, it can be simplified or used to protect the key messages as a part of protocol.
\end{itemize}

Overall, there always exists a trade-off between security, efficiency and utility, in which it is more practical to select proper techniques and privacy levels for specific customer requirements.
\subsection{Enhanced Communication Schemes}\label{subsec:enchanced_comm}
\begin{figure}
\centering
\includegraphics[width=3.5in]{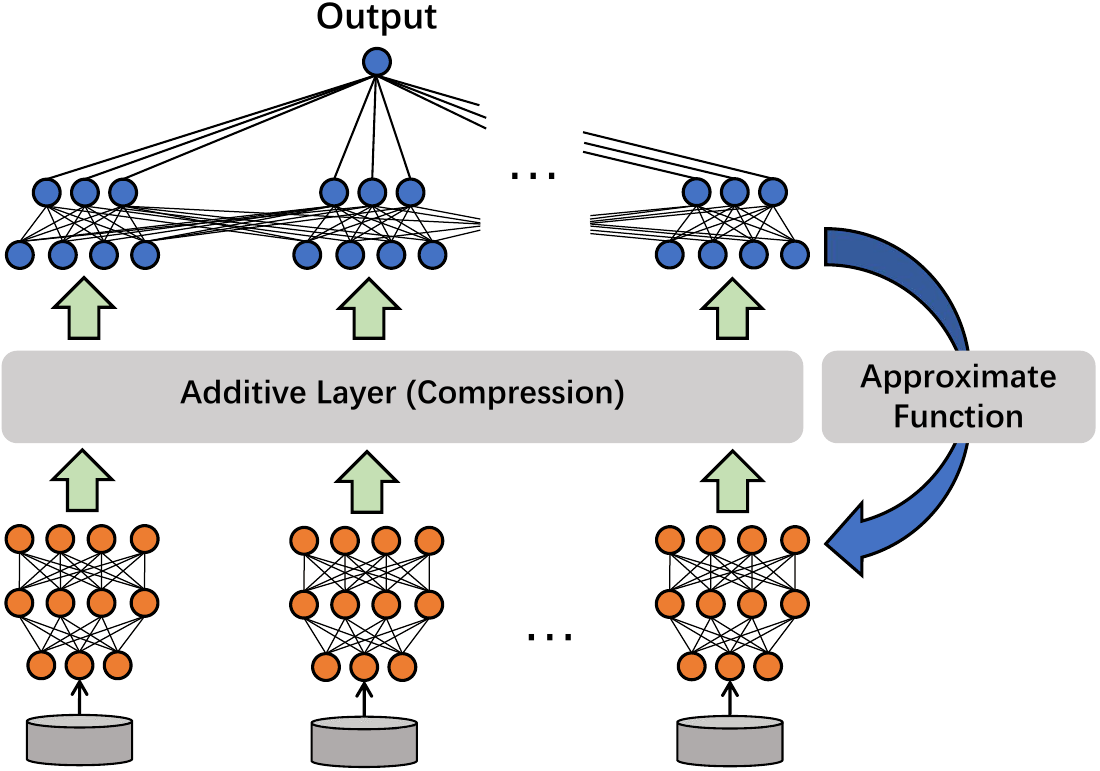}
\caption{A diagram for VFL with the compression function. The compression function can be viewed as an additive layer and it is usually not differentiable, and thus an approximate function can be used to compensate compression errors. }
\label{fig:VFL_additive_layer}
\end{figure}

In this subsection, we will introduce three classic schemes to reduce the communication cost in VFL, i.e., transmission compression, model pruning and data sampling.
\begin{itemize}
\item[$\bullet$]\textbf{Transmission compression.} To reduce the communication cost,  compressing the transmission messages can be an alternative scheme.
Specifically, for one epoch training, VFL requires two transmission processes, i.e., forward and backward transmissions.
For the forward and backward transmissions, we can observe that a large compression ratio will reduce the communication cost much, but involve a large training error, and then affect the updates of bottom and top models.
Unlike HFL, most of classic compression schemes, e.g., uniform quantization,
are not workable since the gradients of forward outputs are unable to derive during back propagation.
Therefore, it will be a promising direction to design an approximate function to assist the backward propagation in VFL as shown in Fig.~\ref{fig:VFL_additive_layer}.
It is also significant to develop the relationship between the quantization level and training performance, which can provide a strong guidance to find a satisfied trade-off between transmission and model performance.
\item[$\bullet$]\textbf{Model pruning.} Model pruning is one of most popular techniques to reduce the computation and transmission resources.
    Specifically, we can drop out some unimportant neurons in the full-connection layer or channels in convolutional layers.
    In this way, the computation and transmission cost will be reduced obviously.
    Therefore, it is significant to further study on evaluating the importance of different neurons or channels, and then achieve the model pruning design with a performance guarantee.
\item[$\bullet$]\textbf{Data sampling.} Data sampling can also be adopted to reduce the communication cost.
In the conventional mini-batch mechanism, all training data will be split into several batches and shuffled, and then VFL would train these batches one by one.
To reduce the communication cost, we can select a part of batches that are important on the model update, via a well designed filtering mechanisms.
Hence, it is a promising direction to design the data selection or combination algorithms to improve the communication efficiency in VFL.
\end{itemize}
\subsection{Novel Mechanisms for Asynchronous VFL}
Participants with heterogenous resources may bring out the asynchrony that requires an intelligent allocator or compensation algorithm to improve the training performance.
Asynchrony has already been studied at many aspects in HFL for stragglers and heterogeneous latency, such as weight design and update compensation for average methods.
In VFL, each participant possesses a unique attribute set, and its forward outputs are much distinct from others.
If certain participants fall behind at one epoch, using the history information (previous outputs) to train this epoch is an alternative.
In addition, in practical scenarios, participants usually have some overlaps in the attribute space, and these overlaps can be explored to address this challenge.
\subsection{Splitting Design}\label{subsec:splitting_des}
The splitting methods for VFL models are key techniques that affect the training performance, communication and computation allocations as well as security and privacy risks.
\begin{itemize}
\item[$\bullet$]\textbf{Effects on communication and computation.}
For one classic learning model, splitting design will directly influence the allocations of training tasks and transmission cost for participants.
In addition, a simple structure of the bottom model, e.g., the linear computation layer, will incur high security and privacy risks and call for further security protections with additional communication and computation costs.
Overall, a personalized design on the cut/splitting layer for different participants and models is required to ensure computation and communication efficiencies.
\item[$\bullet$]\textbf{Effects on security and privacy.}
The splitting rule determines the type of the transmitting messages, which are required to be protected in the learning process.
Involving a privacy-preserving and complexity-acceptability neural network structure for the bottom model may be an alternative.
\item[$\bullet$]\textbf{Effects on model performance.}
Some well-designed learning structures need to extract the relations among original attributes, such as Wide$\&$Deep and DeepFM.
However, VFL requires each participant to keep their raw data locally.
Therefore, we should design the splitting methods while maintaining these specific structures, as well as designing efficient layers to compensate damages.
\end{itemize}
\section{Experiment Results}\label{sec:experi_results}
In this section, we provide experiments to demonstrate the aforementioned issues and discuss some possible solutions.
For each experiment, we first divide the original training data into three parts (three participants) according to the attribute number, and conduct the forward and backward propagations as shown in Fig.~\ref{fig:VFL_systems}.
We evaluate the prototype on the well-known classification dataset Adult and Avazu.
The adult dataset contains $48,842$ records of individuals with $11$ attributes from $1994$ US Census, and is used to predict if an individual's annual income exceeds $50K$ which can been viewed as a binary classification problem.
The Avazu dataset consists of $21$ attributes including $14$ continuous attributes and $7$ categorical attributes, and targets at predicting whether a mobile ID will be clicked.
In addition, we adopt a classic MLP network consisting of three hidden layers with $48$, $96$ and $196$ units, respectively.
The second layer is chosen as the cut/splitting layer, where the size of the forward output of each bottom model will be $32$ and the input size of the top model will be $96$.
We adopt the Adam optimizer and the learning rate is set to $0.0002$.
\subsection{DP assisted VFL}
\begin{figure}[!t]
\centering
\includegraphics[width=3.5in]{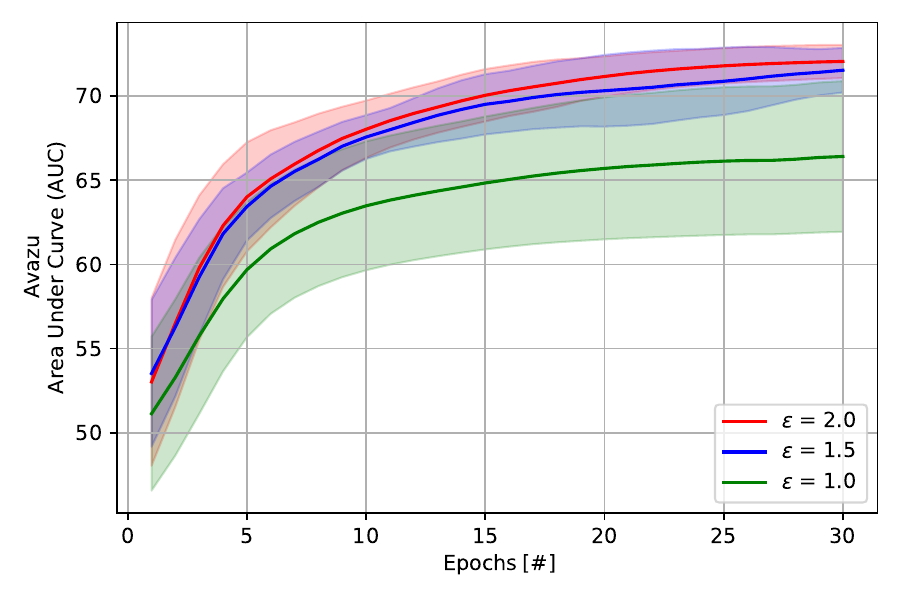}
\caption{AUC with different privacy levels in the VFL framework with three participants (Avazu dataset).}
\label{fig:Avazu_DP_eps}
\end{figure}
First, we evaluate the relationship between the DP mechanism (being against the inference attack) and the test area under curve (AUC) using the Avazu dataset as discussed in~subsection~\ref{subsec:priv_pre}.
In this scenario, Gaussian mechanism has been adopted to preserve the participants' privacy with $\varepsilon = 1$, $1.5$ and $2$, where the noise vector is added on the forward output for each sample.
As shown in Fig.~\ref{fig:Avazu_DP_eps}, we can observe that the AUC performance is largely affected by the added noise.
This is due to the fact that the perturbed forward output will make the loss function value biased.
Therefore, the fundamental relationship between the convergence bound and the privacy level needs to be characterized to achieve configurable trade-off requirements.
\subsection{Compression empowered Communication Efficiency}
\begin{figure}[!t]
\centering
\includegraphics[width=3.5in]{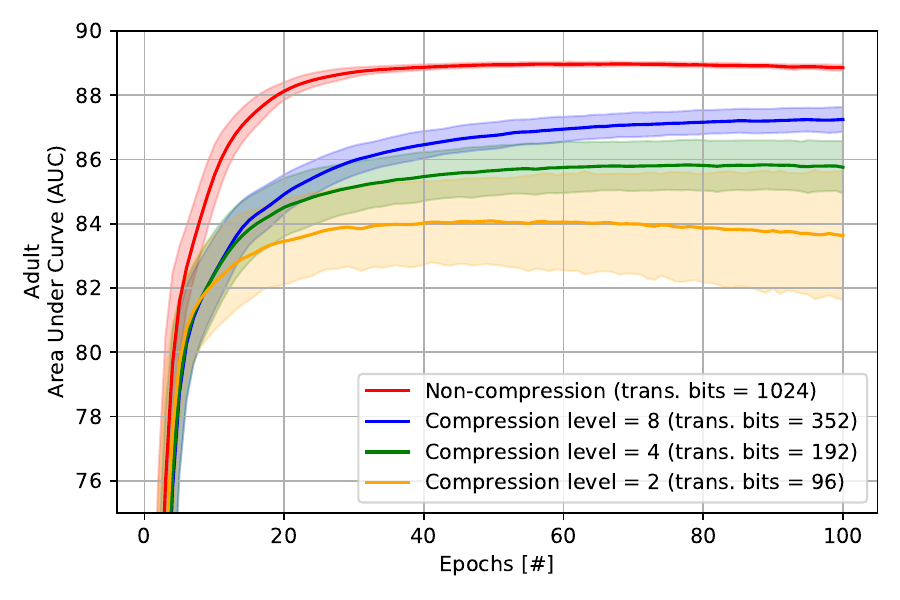}
\caption{The test AUC with different compression levels in the VFL framework (Adult dataset).}
\label{fig:Adult_compression_level}
\end{figure}

\begin{figure}[!t]
\centering
\includegraphics[width=3.5in]{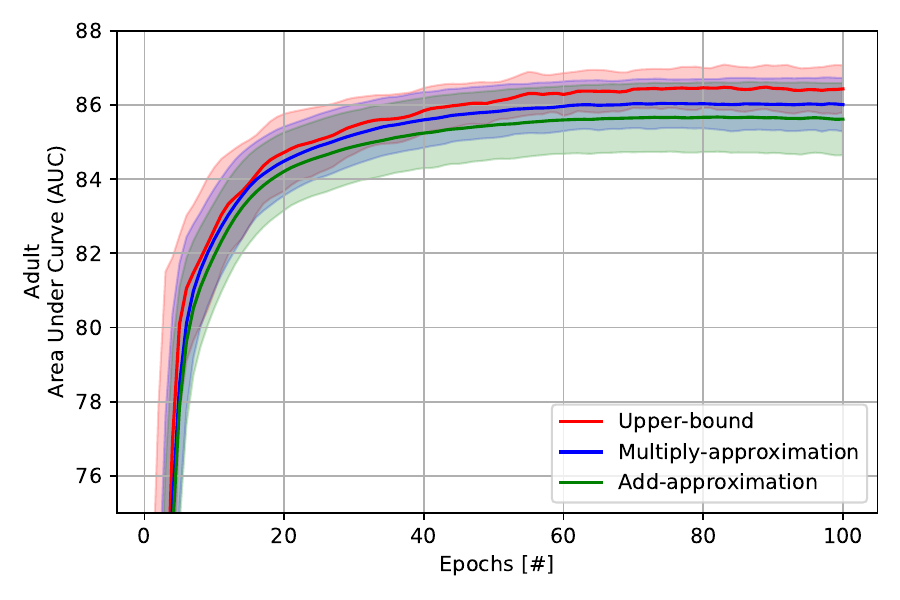}
\caption{The test AUC with different approximation functions under the same compression level in the VFL framework (Adult dataset).}
\label{fig:Adult_compression_approximation}
\end{figure}

To reduce the communication cost, we can compress forward outputs by the well-designed compression method~\cite{Jiang2018SketchML} as mentioned in subsection~\ref{subsec:enchanced_comm}.
In this experiments, we adopt the uniform quantization and the addition approximation for the backward propagation.
Specifically, values in forward outputs will be uniformly divided into $N$ buckets (this number is corresponding to the compression level) using the maximum and minimum values as intervals, and then the mean value of each bucket will be utilized to replace the values located in bucket for forward outputs.
The addition approximation means that when the VFL system derives the gradients of forward outputs, we use $\boldsymbol{o}+\boldsymbol{e}$ to represent the practical compression function, where $\boldsymbol{o}$ is the forward output vector and $\boldsymbol{e}$ is the bias caused by the compression.
The AUC performance of different compression levels, as well as the transmission cost for each level, are shown in Fig.~\ref{fig:Adult_compression_level}.
We can observe that the transmission cost can be reduced obviously, with a high compressing rate while the AUC performance is largely affected.
This is due to the fact that a low compression rate will result in a large bias compared with the original loss function value.

To improve the model performance, we also evaluate the AUC performance with different approximation functions under the same compression level in Fig.~\ref{fig:Adult_compression_approximation}.
The multiply approximation function can be expressed as $\boldsymbol{h}_{i}\boldsymbol{o}_{i}$, where $\boldsymbol{o}_{i}$ is the $i$-th value in $\boldsymbol{o}$ and $\boldsymbol{h}_{i}$ is its corresponding scaling factor, and is comprehensible.
The upper-bound approximation function is designed by the form $\log(a\boldsymbol{o}_{i}+b)+c$, where $a$, $b$ and $c$ can be estimated by the breakpoints in the original compression function.
It can be noticed that the upper-bound and multiplication functions can compensate the bias more caused by the forward compression compared with the addition approximation.
However, it is necessary to explore a more efficient approximation function to improve the model performance under compression.
\subsection{Splitting Design for Resource Allocation}
\begin{figure}[!t]
\centering
\includegraphics[width=3.5in]{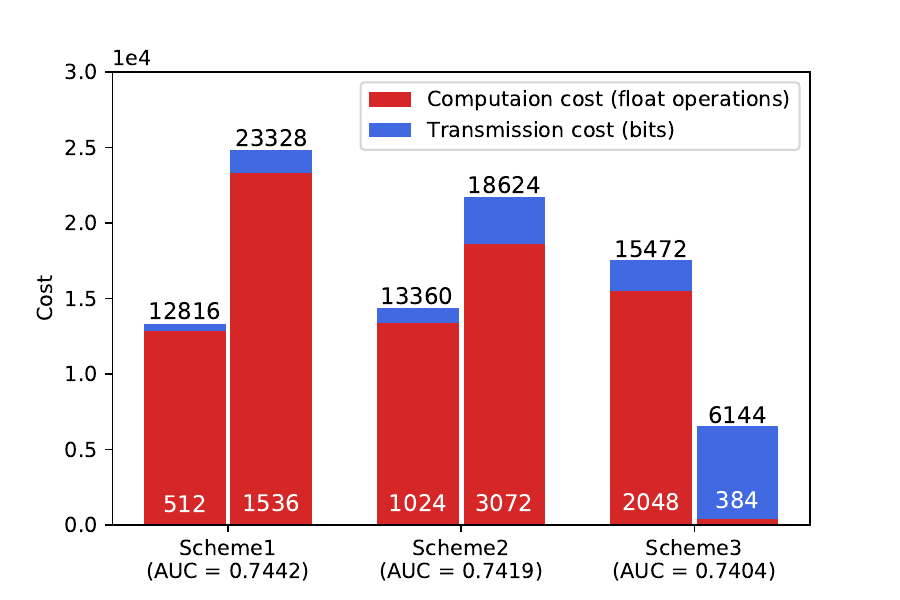}
\caption{Communication and computation cost in the VFL framework with three splitting schemes. The \textbf{left and right bars} of each scheme represent the cost of the \textbf{guest organizations and host organization}, respectively (Avazu dataset).}
\label{fig:Avazu_splitting_cost}
\end{figure}

This subsection will evaluate the effects on the performance, and computation and transmission cost for different splitting methods as discussed in subsection~\ref{subsec:splitting_des}.
We adopt three different splitting designs to train a VFL model with the Avazu dataset as follows:
\begin{itemize}
\item[$\bullet$] \textbf{Scheme 1:} splitting the first hidden layer, and thus each guest will output $16$ values.
\item[$\bullet$] \textbf{Scheme 2:} splitting the second hidden layer, and thus each guest will output $32$ values.
\item[$\bullet$] \textbf{Scheme 3:} splitting the third hidden layer, and thus each guest will output $64$ values.
\end{itemize}
We can notice that the test AUC decreases with deeper cutting layers.
The reason is that when the splitting layer is closer to the raw data, more information of the original data can be combined.
However, the simple data preprocess, e.g., a full-connection layer, makes privacy information protection challenging.
From the view of resource consumes, as shown in Fig.~\ref{fig:Avazu_splitting_cost}, we can observe that when the splitting layer keeps away from the raw data, the computation cost of the guest organization (attribute owner) will be large and the one of the host organization (label owner) will be few.
We can also note that the transmission cost is determined by the unit size of the splitting layer.
If the splitting layer possesses more units, both participants and label owner need to consume a larger transmission cost.
\section{Conclusion}\label{sec:conclusions}

In this article, we have investigated potential challenges and unique issues in VFL from four aspects, i.e., security and privacy risks, expensive computation and communication costs, structural damage and system heterogeneity.
We have pointed out that the splitting design should be adapted to the model's specific structure, and it also affects the privacy and security protection as well as computation and communication efficiency.
In addition, we have discussed possible solutions for the considered issues one by one in designing VFL systems.
Lastly, we have evaluated the studied issues and solutions using two real-world datasets, e.g., DP assisted VFL, compression empowered communication efficiency, and splitting design for resource allocation.


%
%
%
%


\bibliographystyle{IEEEtran}
\bibliography{reference}

\end{document}